\documentclass[10pt,twocolumn,letterpaper]{article}

\usepackage{iccv}              

\usepackage{booktabs}
\usepackage{pifont}

\newcommand{\model}{\textsc{MTU3D}\xspace}
\usepackage{amsmath}
\usepackage{algorithm}
\usepackage{algpseudocode}
\usepackage{multirow}
\usepackage{booktabs}

\usepackage{marvosym}

\newcommand{\cmark}{\ding{51}} 
\newcommand{\xmark}{\ding{55}} 

\definecolor{iccvblue}{rgb}{0.21,0.49,0.74}
\usepackage[pagebackref,breaklinks,colorlinks,allcolors=iccvblue]{hyperref}

\def\paperID{12422} 
\def\confName{ICCV}
\def\confYear{2025}

\title{\textit{Move to Understand a 3D Scene}: Bridging Visual Grounding and Exploration for Efficient and Versatile Embodied Navigation}

\author{
\begin{tabular}{cccccc}
Ziyu Zhu$^{1,2*\dagger}$, 
& Xilin Wang$^{2,4}$, & Yixuan Li$^{2,3}$, & Zhuofan Zhang$^{1,2}$, & Xiaojian Ma$^{2}$, & Yixin Chen$^{2}$,
\end{tabular}
\\
\begin{tabular}{cccccc}
Baoxiong Jia$^{2}$, & Wei Liang$^{3}$, & Qian Yu$^{4}$,  & Zhidong Deng$^{1\dagger}$\Letter, & Siyuan Huang$^2$\Letter, & Qing Li$^{2}$\Letter
\end{tabular} \\
\begin{tabular}{ccc}
$^1$Tsinghua University & $^3$Beijing Institute of Technology  & $^4$Beihang University
\end{tabular} \\
\begin{tabular}{c}
$^2$State Key Laboratory of General Artificial Intelligence, BIGAI, China
\end{tabular} \\
\begin{tabular}{c}
\textbf{Project page: \href{https://mtu3d.github.io}{mtu3d.github.io}}
\end{tabular} \\
\vspace{-15mm}
}

\begin{document}

\twocolumn[{%
\maketitle
\begin{center}
    \centering
    \captionsetup{type=figure}
    \includegraphics[width=0.97\textwidth]{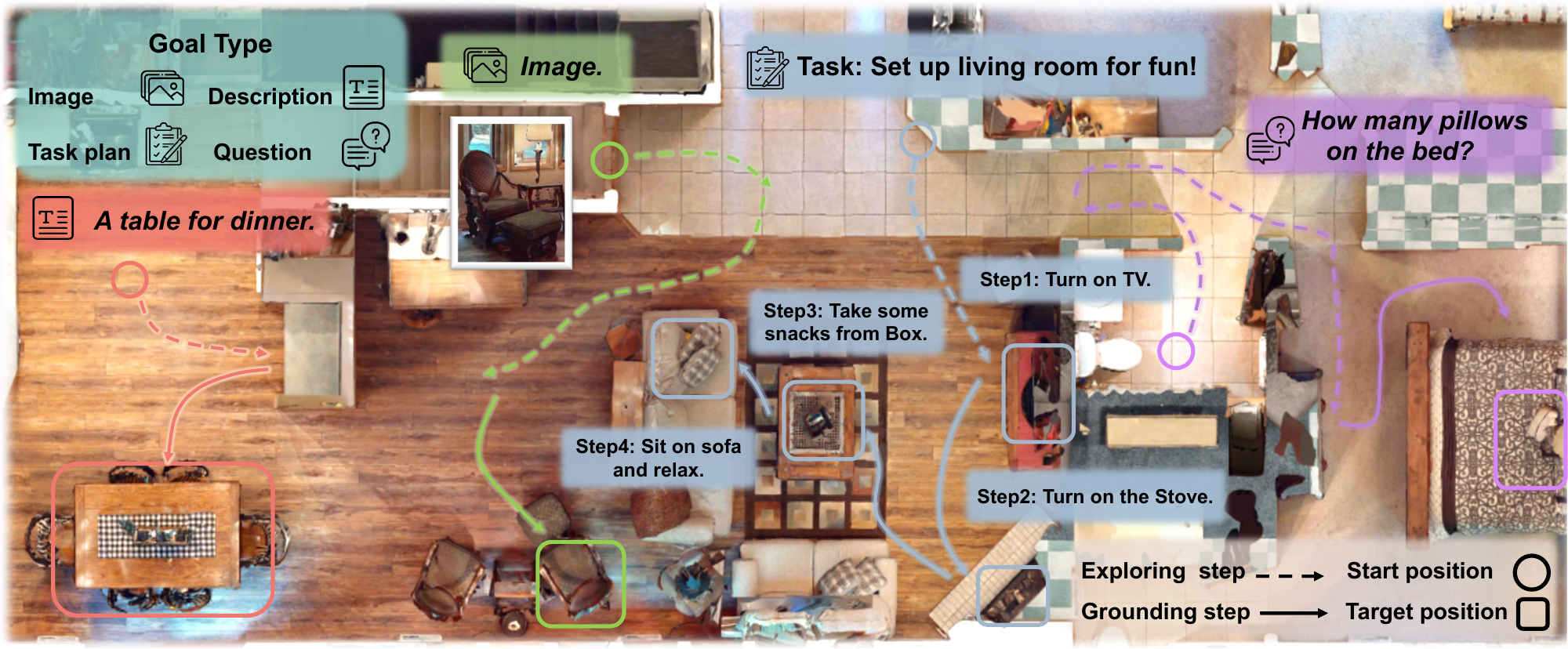}
    \captionof{figure}{\textbf{\model} is a versatile embodied navigation model capable of processing diverse inputs, including \textit{object categories}, \textit{image snapshots}, \textit{natural language descriptions}, \textit{task plan sequences}, and \textit{questions}. It iteratively explores the environment and performs visual grounding to reach the target location. Through large-scale \textbf{V}ision-\textbf{L}anguage-\textbf{E}xploration pre-training, it achieves state-of-the-art performance across various benchmarks including open-vocabulary, multi-modal lifelong, and sequential navigation.}
    \label{fig:teaser}
\end{center}%
}]

\begin{abstract}
Embodied scene understanding requires not only comprehending visual-spatial information that has been observed but also determining where to explore next in the 3D physical world. Existing 3D Vision-Language (3D-VL) models primarily focus on grounding objects in static observations from 3D reconstruction, such as meshes and point clouds, but lack the ability to actively perceive and explore their environment. To address this limitation, we introduce \underline{\textbf{M}}ove \underline{\textbf{t}}o \underline{\textbf{U}}nderstand (\textbf{\model}), a unified framework that integrates active perception with \underline{\textbf{3D}} vision-language learning, enabling embodied agents to effectively explore and understand their environment. This is achieved by three key innovations: 1) Online query-based representation learning, enabling direct spatial memory construction from RGB-D frames, eliminating the need for explicit 3D reconstruction. 2) A unified objective for grounding and exploring, which represents unexplored locations as frontier queries and jointly optimizes object grounding and frontier selection. 3) End-to-end trajectory learning that combines \textbf{V}ision-\textbf{L}anguage-\textbf{E}xploration pre-training over a  million diverse trajectories collected from both simulated and real-world RGB-D sequences. Extensive evaluations across various embodied navigation and question-answering benchmarks show that MTU3D outperforms state-of-the-art reinforcement learning and modular navigation approaches by 14\%, 23\%, 9\%, and 2\% in success rate on HM3D-OVON, GOAT-Bench, SG3D, and A-EQA, respectively. \model's versatility enables navigation using diverse input modalities, including categories, language descriptions, and reference images. The deployment on a real robot demonstrates \model's effectiveness in handling real-world data. These findings highlight the importance of bridging visual grounding and exploration for embodied intelligence.
\end{abstract}    
\section{Introduction}
\label{sec:intro}

Embodied scene understanding requires not only recognizing observed objects but also actively exploring and reasoning in the 3D physical world~\cite{sem-exp,3d-vista,eai-survey,embodied}. Imagine stepping into an unfamiliar room with the goal of \textit{\textbf{``find something to eat''}}. As a human, your instinct would be to explore: perhaps heading to the kitchen first, then scanning the countertops, checking the fridge, or even looking around for a dining area. Your search is driven by a seamless combination of commonsense knowledge, spatial reasoning, and visual grounding~\cite{thinking-in-space,sem-exp,behavior-1k,alfred}. Similarly, an embodied agent navigating a new environment must operate in a continuous closed-loop cycle of exploration, perception, reasoning, and action~\cite{huang2023embodied,alfred,embodiedsam}. A crucial part of this process is understanding both 3D vision and language~\cite{scanrefer,referit3d,multi3drefer,scanents,scanqa,sqa3d,3dgqa} (3D-VL), enabling the agent to think spatially and make informed decisions about where to explore~\cite{pq3d,embodiedlang,teach}.

In recent years, we have seen significant progress in the field of 3D-VL~\cite{peng2023openscene,3d-vista,vil3dref,pq3d,huang2023embodied,3d-llm}. These models leverage 3D reconstructions to perform visual grounding~\cite{scanrefer,viewrefer,embodiedscan,sg3d}, question answering~\cite{scanqa,sqa3d,scanreason}, dense captioning~\cite{scan2cap}, and situated reasoning~\cite{sqa3d,huang2023embodied}. 
Recent approaches, such as 3DVLP~\cite{3d-contrastive}, PQ3D~\cite{pq3d} and LEO~\cite{huang2023embodied}, aim to handle multiple tasks within a single architecture by pre-training~\cite{3d-vista,3d-contrastive} or unified training~\cite{huang2023embodied,3d-llm,3djcg}.

However, existing 3D-VL models rely on static 3D representations~\cite{mask3d,takmaz2023openmask3d, scanrefer}, assuming that a complete reconstruction of the environment is available beforehand~\cite{3d-vista,vil3dref}. While effective for offline vision language grounding, this assumption is impratical for real-world embodied agents, operating in partially observable and dynamic environments~\cite{alfred,openeqa,behavior-1k}. Moreover, these models typically lack active perception and exploration capabilities~\cite{pq3d,hm3d-ovon,goat-bench}. In contrast, reinforcement learning (RL)-based embodied agents can explore environments but often struggle with sample inefficiency~\cite{ver}, poor generalization due to limited training data~\cite{pirlnav,spoc,ppo} and the lack of explicit spatial representation. \textit{\textbf{Bridging passive 3D-VL grounding and active exploration remains a key challenge in developing intelligent systems capable of efficiently exploring and understanding the 3D world.}}

\begin{figure}[!t]  
    \centering
    \includegraphics[width=\linewidth]{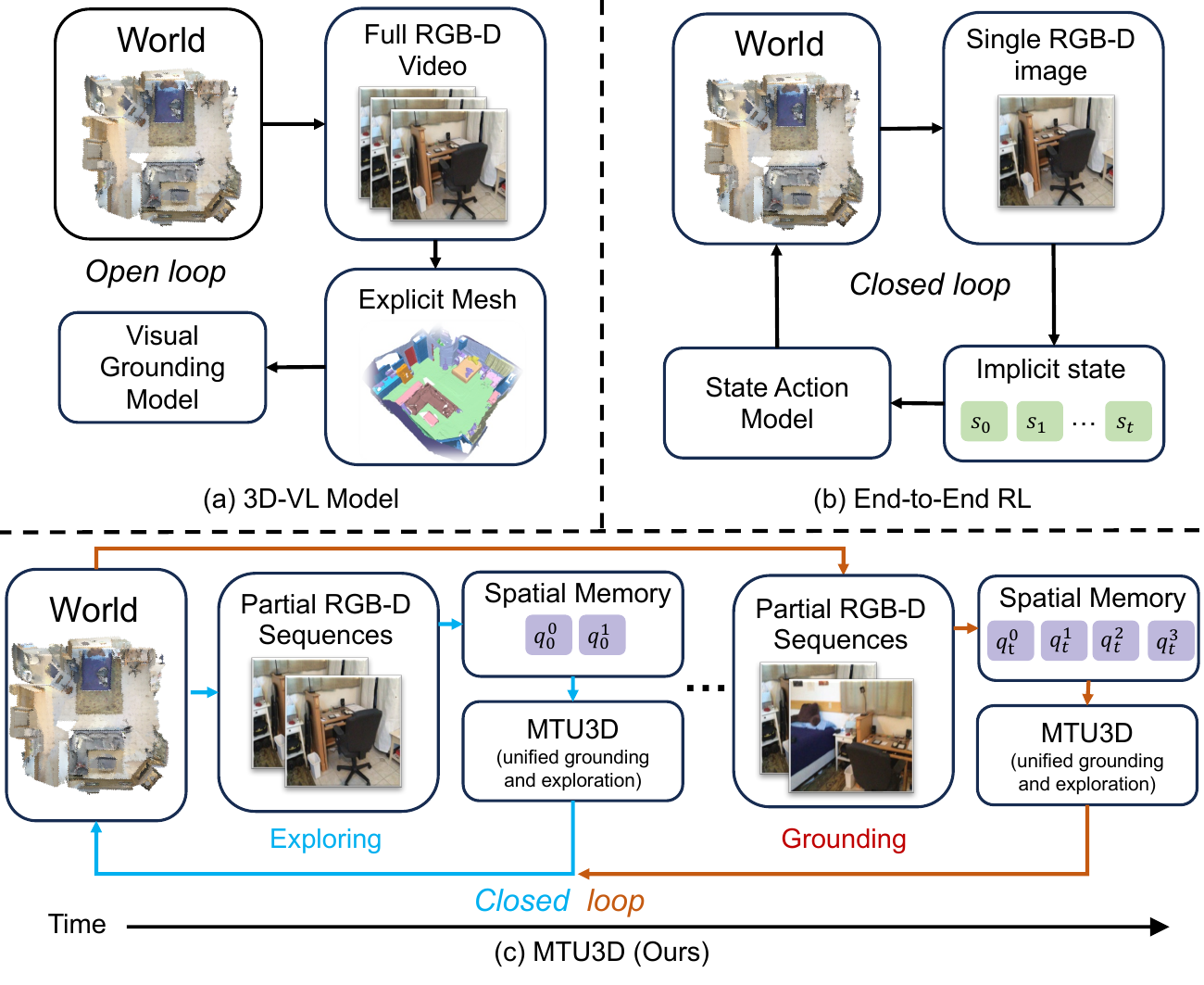} 
    \caption{Our approach bridges online exploration with dynamically spatial memory updates for lifelong grounding.} 
    \label{fig:motivation} 
    \vspace{-7mm}
\end{figure}

\renewcommand{\thefootnote}{} 
\footnotetext{*Work done as an intern at BIGAI.~\Letter~Corresponding author. ~$\dagger$ Department of Computer Science, THUAI, BNRist, Tsinghua University, Beijing 100084, China.}

To develop a 3D-VL model with active perception capabilities, three key challenges must be addressed: First, how to effectively learn online representations from raw RGB-D inputs without costly 3D reconstruction, ensuring rich semantics, spatial awareness, and lifelong memory? Second, the joint optimization of object grounding and spatial exploration remains underexplored. Third, training embodied agents requires large-scale trajectory data for robust exploration strategies, yet collecting diverse real-world trajectories presents significant challenges, and methods for effectively leveraging such data remain an open problem.

To address these challenges, we propose \underline{\textbf{M}}ove \underline{\textbf{to}} \underline{\textbf{U}}nderstand (\textbf{\model}), a unified framework that bridges visual grounding and exploration for versatile embodied navigation as shown in \cref{fig:motivation}. Our approach introduces three key innovations: 1) \textit{Online Query Representation Learning}. Our model processes raw RGB-D frames as input. It generates single-frame local queries and writes them to a global \textit{spatial memory bank}. By leveraging feature extraction and segment priors from 2D foundation models like DINO~\cite{dino,dinov2} and SAM~\cite{SAM}, our query representations capture rich semantics and precise 3D spatial information~\cite{embodiedsam}. 2) \textit{Unified Exploration-Grounding Objective}: We introduce a joint optimization framework where unexplored regions are represented as frontier queries. This allows for simultaneous learning of object grounding and exploration. By feeding object queries retrieved from spatial memory bank along with frontier queries~\cite{fbe} detected from the occupancy map~\cite{real-time-occ}, we enable a cohesive training process that integrates both tasks. 3) \textit{End-to-End Vision-Language-Exploration (VLE) Pre-training}: We train \model using large-scale trajectory data, combining over a million real-world RGB-D trajectories and simulated data from HM3D in total. To enhance training diversity, we develop an automatic trajectory mixing strategy that blends expert and noisy navigation data~\cite{pirlnav}. After VLE pre-training, \model can be seamlessly transferred to both simulated environments and real-world scenarios for inference.

Extensive experiments on embodied navigation and question-answering benchmarks demonstrate that MTU3D outperforms existing offline modular-based and RL-based approaches, achieving higher exploration efficiency, better generalization to unseen environments, and improved real-time decision-making. Specifically, \model improves the state-of-the-art results by 13.7\%, 23.0\%, and 9.1\% in SR, and 2.4\%, 13.0\%, and 6.3\% in SPL on HM3D-OVON~\cite{hm3d-ovon}, GOAT-Bench~\cite{goat-bench}, and SG3D~\cite{sg3d}, respectively. When combined with a large vision-language model, serving as its trajectory generator, our approach improves the embodied question answering for LM-SR by 2.4\% and LLM-SPL by 29.5\%. Furthermore, we deploy the model on a real robot, demonstrating its effectiveness in handling realistic 3D environments. These results highlight the significance of bridging visual grounding and exploration as a crucial step toward efficient, versatile, and generalizable embodied intelligence.

\begin{table}[!t]
    \small
    \centering
    \resizebox{\linewidth}{!}{ 
    \begin{tabular}{lcccc}
        \toprule
        Method & Online & Exploration & Grounding & Lifelong \\
        \midrule
        RL    & \cmark & \cmark & \xmark & \xmark \\
        3D-VL  & \xmark & \xmark & \cmark & \cmark \\  \midrule
        \model (Ours)  & \cmark & \cmark & \cmark & \cmark \\
        \bottomrule
    \end{tabular}
      }
          \vspace{-2mm}
    \caption{Comparison of different paradigms. \model uniquely integrates advantages from both sides, supporting online exploration and lifelong visual grounding. }
    \vspace{-6mm}
    \label{tab:comparison}
\end{table}

Our main contributions can be summarized as follows:
\begin{itemize}[leftmargin=*,noitemsep,topsep=0pt]
    \item We present \textbf{\model}, \textit{bridging visual grounding and exploration} for efficient and versatile embodied navigation.
    \item We propose \textit{a unified objective that jointly optimizes grounding and exploration}, leveraging their complementary nature to enhance overall performance.
    \item We propose a novel \textit{vision-language-exploration training} scheme, leveraging large-scale trajectories from simulation and real-world data. 
    \item Extensive experiments validate the effectiveness of our approach, demonstrating \textit{significant improvements in exploration efficiency and grounding accuracy} across open-vocabulary navigation, multi-modal lifelong navigation, task-oriented sequential navigation, and active embodied question-answering benchmarks.
\end{itemize}
\section{Related work}
\label{sec:related}
\noindent \textbf{3D Vision-Language Understanding.}
In recent years, 3D vision-language (3D-VL) learning~\cite{embodied,embodiedlang,teach,3d-vista,huang2023embodied} has attracted significant attention, with focus on grounding language in 3D scenes by understanding spatial relationships~\cite{referit3d,3rscan}, object semantics~\cite{scanrefer,viewrefer,openseg,takmaz2023openmask3d}, and scene structures~\cite{embodiedscan,jia2024sceneverse,li2022panoptic,concept-fusion}. A wide range of tasks has emerged in this domain, including 3D visual grounding~\cite{scanrefer,referit3d,multi3drefer,scanents}, question answering~\cite{scanqa,sqa3d,3dgqa}, and dense captioning~\cite{scan2cap}. More recently, the field has expanded to cover intention understanding and task-oriented sequential grounding~\cite{sg3d}, further pushing the limits of 3D-VL models in complex reasoning~\cite{saycan, sayplan} and interaction~\cite{voxposer,song2023llm,liang2023code}. Existing 3D-VL models can be broadly categorized into task-specific models~\cite{vil3dref,butd,viewrefer,transrefer3d,yuan2021instancerefer} with specialized architectures for individual tasks, pretrained models~\cite{3d-vista,3d-contrastive} that leverage large-scale multi-modal data to improve generalization, and unified models~\cite{pq3d,huang2023embodied,ll3da,3d-llm,unit3d,3d-vla} that seek to handle multiple tasks within a single framework.  Despite progress, a key limitation of existing 3D-VL models is their reliance on static 3D representations (e.g., precomputed meshes~\cite{mask3d,takmaz2023openmask3d} or point clouds~\cite{scanrefer}), making them unsuitable for real-world embodied AI (EAI), where agents must explore and perceive environments in real-time. EmbodiedSAM~\cite{embodiedsam} partly addresses this issue by taking streaming RGB-D video as input for online 3D instance segmentation~\cite{scenegraph,concept-graph}; however, it lacks active exploration and high-level reasoning capability. In contrast, our \model framework is proposed as a unified model aiming to simultaneously learn scene representations, exploration strategies, and grounding directly from \textit{dynamic spatial memory bank} during online RGB-D exploration.

\noindent \textbf{Embodied Navigation and Reasoning.}
The recent advances in embodied AI, particularly embodied navigation~\cite{szot2021habitat, hong2021vln,realfred,habitat} and reasoning~\cite{das2018embodied, alfred, majumdar2024openeqa,alfred,behavior-1k,ren2024explore}, primarily rely on three crucial capabilities: perception, reasoning, and exploration.
Several benchmarks are proposed to assess navigation capabilities~\cite{goat-bench,hm3d-ovon} across different goal specifications (images, objects, or language instructions), examine the sequential awareness~\cite{sg3d}, or question-answering in an embodied setting~\cite{openeqa}. Efforts to tackle these challenges generally follow two main approaches~\cite{uni-navid,tango,goat-bench,lin2024navcot}: end-to-end reinforcement learning and modular architectures. End-to-end approaches, like PIRLNav~\cite{pirlnav} and VER~\cite{ver}, utilize RNNs or transformers~\cite{poliformer,one-ring} trained directly for navigation tasks, integrating perception and reasoning to take actions. However, its end-to-end nature without explicit 3D representations limits its performance under complex instructions in intricate environments~\cite{yu2023frontier,hong2021vln}. In contrast, modular approaches~\cite{objectgoalnav,yu2023frontier,yokoyama2024vlfm,ren2024explore,li2023blip,3D-mem} decompose navigation into specialized components, maintaining distinct models for mapping and navigation policies.
More recently, CLIP on Wheels~\cite{gadre2023cows,clip,uniclip} leverages pre-trained vision-language models to interpret navigation goals without fine-tuning~\cite{zson}. 
Unlike these methods as in \cref{tab:comparison}, our model employs a vision-language-exploration training paradigm, actively exploring to construct a scene representation and reasoning within an end-to-end system.

\section{\model}
\begin{figure*}[!t]  
    \centering
    \includegraphics[width=1\textwidth]{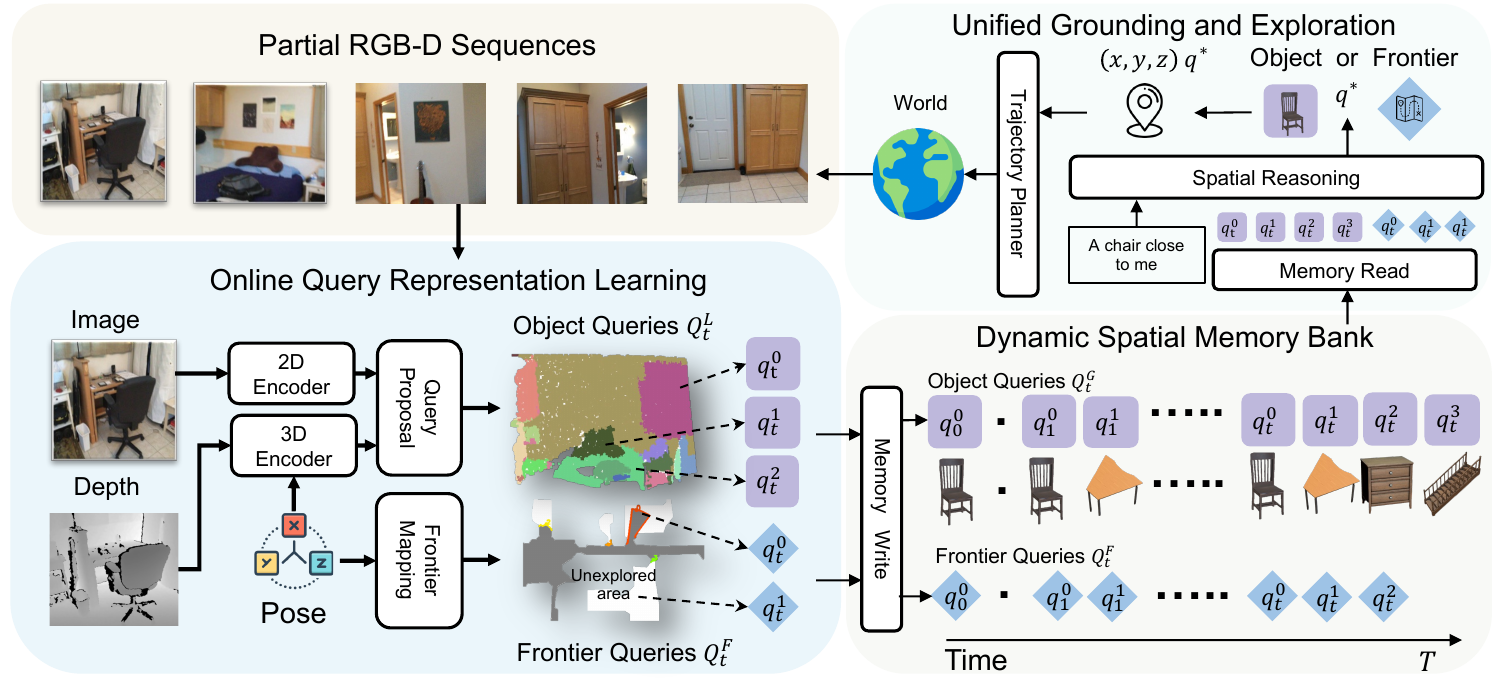} 
    \caption{Our proposed model processes RGB-D sequences to generate object queries, which are stored in a memory bank. The spatial reasoning layer then selects either object queries or frontier points for exploration. These selected query locations are passed to a trajectory planner that guides navigation and facilitates the acquisition of new RGB-D sequences, creating a continuous perception-action loop. } 
    \label{fig:model} 
\end{figure*}

In this section, we present the architecture of our model in \cref{fig:model} and the training pipeline. We begin by detailing our approach to query representation learning, which extracts object and frontier queries from partial RGB-D sequences and dynamically stores them in a spatial memory bank. Next, we introduce our unified grounding and exploration objective, where queries are processed through a spatial reasoning layer for selection. Finally, we describe our trajectory collection strategy and training procedure.
\label{sec:method}

\subsection{Online Query Representation Learning}
Our model takes as input a partial RGBD sequence spanning an arbitrary time range, denoted as $O = [o_{t_1}, o_{t_2}, \dots, o_{t}]$. At each timestep $t$, we extract local queries $Q_t^L$, which are subsequently aggregated across the interval $t_1$ to $t_2$ to generate global queries $Q_{t_2}^G$ that are stored in a memory bank. This process operates in an online manner, enabling flexible processing at any timestep. Our online representation learning framework comprises three essential components, as illustrated below.  

\noindent \textbf{2D and 3D Encoding.}
For each input observation $o_t = [I_t, D_t, P_t]$, we process three components: RGB image $I_t \in \mathbb{R}^{H \times W \times 3}$, depth image $D_t \in \mathbb{R}^{H \times W}$, and camera pose $P_t \in SE(3)$. Following ESAM~\cite{embodiedsam}, we apply FastSAM~\cite{SAM,fastsam} to segment the image into distinct regions, generating a segment index map $S_t \in \mathbb{N}^{H \times W}$ where each value represents a region ID. For 2D feature extraction, we process $I_t$ through a DINO backbone~\cite{dino,dinov2} to obtain pixel-level features $F_t^{2D} \in \mathbb{R}^{H \times W \times C}$. These features are pooled according to $S_t$, producing segment-level representations~\cite{pq3d} $\hat{F}_t^{2D} \in \mathbb{R}^{M \times C}$, with $M$ denoting the number of segments identified by FastSAM. Each segment corresponds to a coarse image region associated with a group of pixels and their respective 3D points.

For 3D feature extraction, we project the depth image $D_t$ into a point cloud and uniformly downsample it to $N$ points $\mathcal{P}_t \in \mathbb{R}^{N \times 3}$, which is processed through a sparse convolutional U-Net~\cite{minkowski,mask3d,mask2former} to generate 3D features $F_t^{3D} \in \mathbb{R}^{N \times C}$, with $C$ being the feature dimension. We then perform segment-wise pooling~\cite{graph-seg} using $S_t$, resulting in segment-level 3D representations $\hat{F}_t^{3D} \in \mathbb{R}^{M \times C}$.

\noindent \textbf{Local Query Proposal.}
After obtaining 2D and 3D features, we generate object queries through a multi-layer perceptron: $Q_t = \mathrm{MLP}(\hat{F}_t^{2D}, \hat{F}_t^{3D}) \in \mathbb{R}^{M \times C}$. These initial queries are refined via PQ3D-inspired~\cite{pq3d,xdecoder} decoder layers, producing local output queries $Q_t^{L}$. Each refined local query $q_t^{L} \in Q_t^{L}$ comprises multiple components: $q_t^{L} = [b_t, m_t, f_t, v_t, s_t]$, where $b_t \in \mathbb{R}^6$ encodes a 3D bounding box in global coordinates, $m_t \in \mathbb{R}^{M}$ represents segment-level mask predictions for instance segmentation, $v_t \in \mathbb{R}^{C}$ captures open-vocabulary feature embeddings for semantic alignment, $f_t \in \mathbb{R}^{C}$ contains output features from the query decoder, and $s_t \in \mathbb{R}$ denotes a confidence score indicating detection reliability.  Each mask in shape $M$ can be converted to a 2D mask ($H \times W$) or a single frame 3D point cloud mask ($N$ points).  As the input point cloud is transformed into world coordinates using pose information, the predicted $b_t$ exists in a shared coordinate system, enabling subsequent query merging operations.

\noindent \textbf{Dynamic Spatial Memory Bank.} We merge local queries with historical queries from our spatial memory bank~\cite{embodiedsam,spatial-mem,dual-memory} by calculating bounding box IoU between current local queries $Q_t^{L}$ and previous global queries $Q_{t-1}^{G}$, generating updated global queries $Q_t^{G}$, which contains $f_t$ and $v_t$ for further grounding and exploration. For matched query pairs, we employ a fusion strategy where bounding box parameters $b_t$, feature embeddings $f_t$, semantic vectors $v_t$, and confidence scores $s_t$ are combined using exponential moving averages, while instance segmentation masks $m_t$ are fused through a union operation following~\cite{embodiedsam}. To identify unexplored regions, we maintain an occupancy map $\mathcal{M} \in \mathbb{R}^{X \times Y \times 3}$ that categorizes spatial segments as occupied, unoccupied, or unknown. Central to our exploration approach is the concept of frontiers, denoted as $Q_{t}^{F}$, which represent boundaries between explored (known) and unexplored (unknown) areas. These frontiers are identified by traversing the boundaries of the explored space and detecting adjacent unknown areas. Each element of $Q_{t}^{F}$ corresponds to a coordinate in 3D space $\mathbb{R}^3$, serving as a potential exploration target~\cite{fbe}. These frontier waypoints are periodically recalculated upon reaching each target position, directing autonomous agents to explore unmapped regions.

\subsection{Unified Grounding and Exploration}
Our unified approach combines object grounding and exploration in a single decision framework. Given a natural language goal $L$ (e.g., ``find a red chair"), our model decides between grounding an object from current global queries $Q_t^G$ or selecting a frontier from $Q_t^F$ for exploration.A spatial reasoning transformer, described in detail in the appendix, integrates language instructions, object representations, and frontier information to produce a unified score $S_t^U = f(Q_t^G, Q_t^F, L)$ for each candidate decision. Frontier points are encoded through a simple two-layer MLP before being fed into the transformer. The input for $Q_t^G$ consists of $f_t$ and $v_t$. We incorporate type embeddings to distinguish between object and frontier queries. For language goals, we use a CLIP text encoder~\cite{clip}, while image-based goals employ the CLIP image encoder~\cite{clipbert,clip}. These embeddings are projected into the transformer's feature space for cross-attention with query representations. The final decision selects the query with the highest score: $q^* = \arg\max_{q_i \in Q_t^G \cup Q_t^F} S_t^U(q_i)$. If $q^* \in Q_t^G$, the system grounds the corresponding object; if $q^* \in Q_t^F$, it navigates to that frontier location. We employ the Habitat-Sim shortest path planner to generate local trajectories. This mechanism dynamically balances exploration and grounding based on current scene knowledge and goal~\cite{sem-exp}. 

\subsection{Trajectory Data Collection}
Our unified model training requires diverse trajectory data, which is difficult to collect manually~\cite{pirlnav,dd-ppo}. We implement a systematic collection process spanning simulated and real environments, combining \textit{visual grounding data} and \textit{exploration data} as shown in \cref{tab:data_sources}.

\noindent \textbf{Visual Grounding Trajectory.}
Our visual grounding data follows the structure (Object $Q_t^G$, Language $L$) $\longrightarrow$ (Decision $S_t^U$). Leveraging offline RGB-D videos in ScanNet~\cite{scannet200,dai2017scannet} as trajectories, we can directly use these paired examples for pre-training. Each sample associates object queries with language descriptions to generate decisions.

\noindent \textbf{Exploration Trajectory.}
Exploration data follows the format (Object $Q_t^G$, Frontier $Q_t^F$, Goal $G$) $\longrightarrow$ (Decision $S_t^U$), with frontiers changing during exploration~\cite{fbe}. Training with only optimal frontiers (closest to target) leads to overfitting. We address this by implementing random frontier selection and a hybrid strategy combining random and optimal approaches~\cite{pirlnav}. We collect trajectories from simulation scans (HM3D~\cite{hm3d-sem} via Habitat-Sim~\cite{habitat}) to apply these varied exploration strategies. Exploration succeeds when the target becomes visible and reachable. To prevent unnecessary exploration, we maintain a visited frontier list and only explore when better potential frontiers exist (much closer to target); otherwise, we raise an exception and terminate the current collection. The complete collection process appears in the appendix.
\begin{table}[!t]
    \centering
    \renewcommand{\arraystretch}{1.2}
    \resizebox{\linewidth}{!}{ 
    \begin{tabular}{lccccccccc}
        \toprule
        \textbf{Source} & \textbf{Scan} & \textbf{Sim} & \textbf{Real} & \textbf{VG} & \textbf{Exp}  & \textbf{Traj} & \textbf{Dec} & \textbf{Goal}  \\
        \midrule
        ScanRefer~\cite{scanrefer} & ScanNet & \xmark & \cmark & \cmark & \xmark & 1202 & 37k & 37k \\
        ScanQA~\cite{scanqa} & ScanNet & \xmark & \cmark & \cmark & \xmark & 1202 & 26k & 30k\\
        Multi3DRefer~\cite{multi3drefer} & ScanNet & \xmark & \cmark & \cmark & \xmark & 1202 & 44k & 44k \\
        SG3D-VG-HM3D~\cite{sg3d} & HM3D & \cmark & \xmark & \cmark & \xmark & 145 & 30k & 30k \\
        SG3D-VG-ScanNet~\cite{sg3d} & ScanNet & \xmark & \cmark & \cmark & \xmark & 1202 & 13k & 13k\\
        Nr3D~\cite{referit3d} & ScanNet & \xmark & \cmark & \cmark & \xmark & 1202 & 30k & 30k \\
        SceneVerse-HM3D~\cite{jia2024sceneverse} & HM3D & \cmark & \xmark & \cmark & \xmark & 145 & 48k & 48k\\
        HM3D-OVON~\cite{hm3d-ovon} & HM3D & \cmark & \xmark & \xmark & \cmark & 290k & 1940k & 290k\\
        GOAT-Bench~\cite{goat-bench} & HM3D & \cmark & \xmark & \xmark & \cmark & 680k & 14119k & 5098k\\
        SG3D-Nav~\cite{sg3d} & HM3D & \cmark & \xmark & \xmark & \cmark & 2k & 34k & 11k\\
        \bottomrule
    \end{tabular}
    }
    \caption{Data source statistic for Vision-Language-Exploration Pre-training. \textbf{Sim} denotes simulation, \textbf{VG} denotes visual grounding, \textbf{Exp} denotes exploration, \textbf{Traj} denotes trajectory, \textbf{Dec} denotes decision and \textbf{Goal} denotes number of goals. }
    \label{tab:data_sources}
\end{table}

\subsection{Vision-Language-Exploration Training}
\noindent \textbf{Stage 1: Low-level Perception Training.}
We utilize RGB-D trajectories from ScanNet and HM3D to train query representation with instance segmentation loss. Our loss function combines multiple components: $\mathcal{L} = \lambda_b \mathcal{L}{\text{box}} + \lambda_m \mathcal{L}{\text{mask}} + \lambda_v \mathcal{L}{\text{vocab}} + \lambda_s \mathcal{L}{\text{score}}$. $\mathcal{L}{\text{box}}$ is the 3D box IoU loss; $\mathcal{L}{\text{mask}}$ and $\mathcal{L}{\text{score}}$ are binary cross-entropy losses; $\mathcal{L}{\text{vocab}}$ is cosine similarity loss. This approach ensures that output local queries $Q_t^L$ effectively capture spatial, semantic, and confidence information.

\noindent \textbf{Stage 2: Vision-Language-Exploration Pre-training.}
In VLE pre-training, we utilize stage 1 output queries to jointly train exploration and grounding. Using the dataset in \cref{tab:data_sources}, we train our decision model on over one million trajectories. The unified decision scores $S_t^U$ are optimized with binary cross-entropy loss, teaching the model to assign higher scores to appropriate query locations based on the current state and goal.

\noindent \textbf{Stage 3: Task-Specific Navigation Fine-tuning.}
During this stage, we employ the same objective as stage 2 and fine-tune our \model to specific navigation trajectories, optimizing performance for targeted deployment scenarios.
\newpage
\section{Experiment}
\subsection{Experimental setting}
\noindent \textbf{Dataset and Benchmarks.}
We evaluate on diverse benchmarks: GOAT-Bench~\cite{goat-bench} (multi-modal lifelong navigation), HM3D-OVON~\cite{hm3d-ovon} (open-vocabulary navigation), SG3D~\cite{sg3d} (sequential task navigation), and A-EQA~\cite{openeqa} (embodied question answering). Common metrics include Success Rate ($\text{\textit{SR}} = \frac{N_\text{success}}{N_\text{total}}$) and Success weighted by Path Length ($\text{\textit{SPL}} = \frac{1}{N_\text{total}} \sum_{i=1}^{N_\text{total}} S_i \cdot \frac{l_i}{\max(p_i, l_i)}$), where $S_i$ indicates success, $l_i$ is shortest path length, and $p_i$ is agent path length. SG3D uses task SR (\textit{t-SR}) to measure step coherence. A-EQA employs LLM match score (\textit{LLM-SR}) and score averaged by exploration length (\textit{LLM-SPL}). We compare against a diverse range of baseline methods, including modular approaches such as GOAT~\cite{goat-bench} and VLFM~\cite{yokoyama2024vlfm}, end-to-end reinforcement learning approaches~\cite{hm3d-ovon, pirlnav}, and video-based approaches~\cite{uni-navid}.

\noindent \textbf{Implementation Details.}
In Stage 1, we train for 50 epochs using AdamW (learning rate 1e-4, $\beta_1$ = 0.9, $\beta_2$ = 0.98) with loss weights $\lambda_b=1.0, \lambda_m=1.0, \lambda_v=1.0, \lambda_s=0.5$. Stages 2 and 3 use identical optimizer settings for 10 epochs each. Both Stages 1 and 2 use 4 transformer layers.  Query proposal is trained in stage 1 then frozen, and spatial reasoning is trained in later stages. All training runs on four NVIDIA A100 GPUs around 164 GPU hours. For simulation evaluation, we follow~\cite{goat-bench,hm3d-ovon,sg3d} using Stretch embodiment (1.41m tall, 17cm base radius), processing 360×640 RGB images $I_t$, depth maps $D_t$, and pose $P_t$ with actions: \texttt{MOVE\_FORWARD}(0.25m), \texttt{TURN\_LEFT}, \texttt{TURN\_RIGHT}, \texttt{LOOK\_UP}, and \texttt{LOOK\_DOWN}. Spatial reasoning is activated upon arrival at each target position. We subsample 18 frames along the trajectory between consecutive target positions.For A-EQA~\cite{openeqa}, our model is used solely to generate the exploration trajectory and collect the corresponding video for each question. 

\subsection{Quantitative Results}
\noindent \textbf{Open Vocabulary Navigation.}
The results in \cref{tab:ovon-result} demonstrate that our proposed MTU3D significantly outperforms all baselines in terms of SR across both Val Seen and Val Unseen settings. Notably, MTU3D achieves the highest SR in Val Unseen (40.8\%), showcasing its strong generalization ability to unseen episodes. This highlights the effectiveness of our approach in handling unseen scenarios compared to prior methods such as RL and behavior cloning (BC), which are not pre-trained on large-scale language data. However, we observe that MTU3D's SPL is lower than Uni-Navid, especially in the Val Synonyms and Val Unseen settings. Given that the HM3D-OVON dataset consists of relatively short trajectories, video-based models inherently gain an advantage because they can directly navigate to the goal upon recognizing a target.

\begin{table}[!h]
\centering
\resizebox{\columnwidth}{!}{ 
\begin{tabular}{lcccccc}
\toprule
 & \multicolumn{2}{c}{\multirow{2}{*}{
 Val Seen}} & \multicolumn{2}{c}{Val Seen} & \multicolumn{2}{c}{\multirow{2}{*}{Val Unseen}} \\
 & & & \multicolumn{2}{c}{Synonyms} & & \\
\cmidrule(lr){2-3} \cmidrule(lr){4-5} \cmidrule(lr){6-7}
Method & SR ($\mathbf{\uparrow}$) & SPL ($\mathbf{\uparrow}$) & SR ($\mathbf{\uparrow}$) & SPL ($\mathbf{\uparrow}$) & SR ($\mathbf{\uparrow}$) & SPL ($\mathbf{\uparrow}$) \\ \hline
BC~ & $11.1$ & $4.5$ & $9.9$ & $3.8$ & $5.4$ & $1.9$ \\
DAgger & $18.1$ & $9.4$ & $15.0$ & $7.4$ & $10.2$ & $4.7$ \\
RL & $39.2$ & $18.7$ & $27.8$ & $11.7$ & $18.6$ & $7.5$ \\
BCRL & $20.2$ & $8.2$ & $15.2$ & $5.3$ & $8.0$ & $2.8$ \\
DAgRL & $41.3$ & $21.2$ & $29.4$ & $14.4$ & $18.3$ & $7.9$ \\
VLFM  & $35.2$ & $18.6$ & $32.4$ & $17.3$ & $35.2$ & $19.6$ \\
Uni-NaVid & $41.3$ & $21.1$ & $43.9$ & $\mathbf{21.8}$ & $39.5$ & $19.8$ \\
TANGO   & $-$ & $-$ & $-$ & $-$ & $35.5$ & $19.5$ \\ 
DAgRL+OD & $38.5$ & $21.1$ & $ 39.0$ & $21.4$ & $ 37.1$ & $\mathbf{19.9}$ \\ \midrule
 \model(Ours) & $\mathbf{55.0}$ & $\mathbf{23.6}$ & $\mathbf{45.0}$ & $14.7$ & $\mathbf{40.8}$ & $12.1$ \\ 
\bottomrule
\end{tabular}
}
    \caption{Open-vocab navigation results on HM3D-OVON~\cite{hm3d-ovon}.}
    \label{tab:ovon-result}
\end{table}

\noindent \textbf{Task-oriented Sequential Navigation.}
\cref{tab:sg3d-result} presents results on task-oriented sequential navigation, a challenging task requiring an embodied agent to understand relationships between task steps. The results show that MTU3D achieves the highest s-SR (23.8\%), t-SR (8.0\%), and SPL (16.5\%) on the SG3D benchmark, demonstrating its effectiveness in sequential task execution and task understanding. Unlike other benchmarks, SG3D emphasizes task consistency across multiple steps, making it more complex. While MTU3D significantly outperforms Embodied Video Agent~\cite{embodied-video-agent} and SenseAct-NN Monolithic~\cite{goat-bench,sg3d}, overall success rates remain lower than in GOAT-Bench and HM3D-OVON, highlighting SG3D’s inherent difficulty in requiring both navigation and sustained task accuracy.


\begin{table}[!h]
    \centering
    \renewcommand{\arraystretch}{1.2}
    \resizebox{0.8\columnwidth}{!}{ 
    \begin{tabular}{lccc}
        \toprule
        & s-SR(↑) & t-SR(↑) & SPL(↑) \\
        \midrule
        Embodied Video Agent & 14.7 & 3.8 & 10.2 \\
        SenseAct-NN Monolithic & 12.1 & 7.7 & 10.1 \\ \midrule
        \model(Ours) & \textbf{23.8} & \textbf{8.0} & \textbf{16.5} \\
        \bottomrule
    \end{tabular}
    }
    \caption{Sequential task navigation results on SG3D-Nav~\cite{sg3d}.}
    \label{tab:sg3d-result}
\end{table}

\begin{table*}[!t]
    \centering
    \small
    \renewcommand{\arraystretch}{1.2}
    \begin{tabular}{lcccccc}
        \toprule
        & \multicolumn{2}{c}{Val Seen} & \multicolumn{2}{c}{Val Seen Synonyms} & \multicolumn{2}{c}{Val Unseen} \\
        \cmidrule(lr){2-3} \cmidrule(lr){4-5} \cmidrule(lr){6-7}
        Method & SR (↑) & SPL (↑) & SR (↑) & SPL (↑) & SR (↑) & SPL (↑) \\ \midrule
        Modular GOAT  & 26.3 & 17.5 & 33.8 & 24.4 & 24.9 & 17.2 \\
        Modular CLIP on Wheels  & 14.8 & 8.7 & 18.5 & 11.5 & 16.1 & 10.4 \\
        SenseAct-NN Skill Chain & 29.2 & 12.8 & 38.2 & 15.2 & 29.5 & 11.3 \\
        SenseAct-NN Monolithic & 16.8 & 9.4 & 18.5 & 10.1 & 12.3 & 6.8 \\ 
        TANGO & - & - & - & - & 32.1 & 16.5 \\
        \midrule
        \model(ours) & \textbf{52.2} & \textbf{30.5} & \textbf{48.4} & \textbf{30.3} & \textbf{47.2} & \textbf{27.7} \\ 
        \bottomrule
    \end{tabular}
    \caption{Multi-modal lifelong navigation results on GOAT-Bench~\cite{goat-bench}.}
    \label{tab:goat-result}
\end{table*}

\noindent \textbf{Multi-modal Lifelong Navigation.} The results in \cref{tab:goat-result} highlight the significant performance improvement of our MTU3D over baseline methods in lifelong setting. Notably, MTU3D achieves the highest SR across all settings, with 52.2\% in Val Seen, 48.4\% in Val Seen Synonyms, and 47.2\% in Val Unseen, demonstrating its superior navigation capability. Compared to open-vocabulary navigation, multi-modal lifelong navigation is a more challenging task due to the need for continuous spatial memory and long-term reasoning. Our model's lifelong spatial memory allows it to retain and utilize past experiences more effectively, leading to a much larger performance gain over baselines. 
Additionally, MTU3D achieves the highest SPL across all settings, with 30.5\% in Val Seen and 27.7\% in Val Unseen, indicating that it not only reaches the goal more accurately but also follows more efficient trajectories. This suggests that our approach effectively balances success rate and efficiency, a crucial factor in lifelong navigation.  Overall, these results emphasize that lifelong spatial memory is key to multi-modal navigation, and MTU3D's substantial improvements validate its effectiveness in handling this complex task.

\noindent \textbf{Active Embodied Question Answering.}
\cref{tab:a-eqa} demonstrate that our MTU3D-enhanced GPT-4V significantly outperforms the baseline GPT-4V model, achieving 44.2\% LLM-SR vs. 41.8\% and a much higher LLM-SPL of 37.0\% vs. 7.5\%. This indicates that our approach enables a more efficient trajectory, avoiding the exhaustive search across all locations that baseline models rely on. Furthermore, GPT-4o with MTU3D achieves even better performance, reaching 51.1\% LLM-SR and 42.6\% LLM-SPL. 

\begin{table}[!h]
    \centering
    \small
    \begin{tabular}{lcc}
        \toprule
        Method & LLM-SR(↑) & LLM-SPL(↑) \\
        \midrule
         \textit{Blind LLMs} &  &  \\
        GPT-4 & 35.5 & N/A \\
        LLaMA-2 & 29.0 & N/A \\
        \midrule
         \textit{LLM with captions} &  &  \\
        GPT-4 w/ LLaVA-1.5 & 38.1 & 7.0 \\
        LLaMA-2 w/ LLaVA-1.5 & 30.9 & 5.9 \\
        \midrule
        \textit{VLMs} &  &  \\
        GPT-4V & 41.8 & 7.5 \\ 
        \midrule
        \multicolumn{3}{l}{\textit{VLMs with MTU3D Trajectory}} \\
        GPT-4V (Ours) & 44.2 & 37.0 \\ 
        GPT-4o (Ours) & \textbf{51.1} & \textbf{42.6} \\ 
        \bottomrule
    \end{tabular}
    \caption{Embodied question answering results on A-EQA~\cite{openeqa}.}
    \label{tab:a-eqa}
\end{table}

\subsection{Discussions}
\begin{figure*}[!t]  
    \centering
    \begin{subfigure}[b]{0.26\textwidth}
        \centering
        \includegraphics[width=\textwidth]{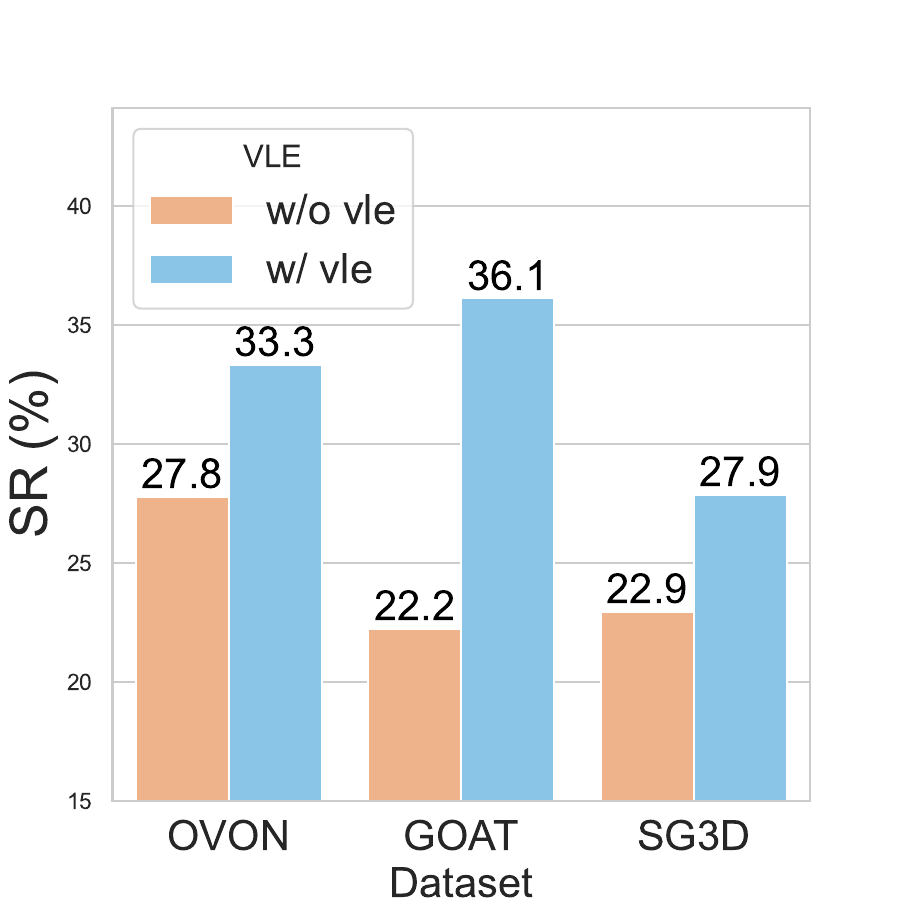}
        \caption{Effect of VLE.}
        \label{fig:pretrain-ablation}
    \end{subfigure}
    \hfill
    \begin{subfigure}[b]{0.25\textwidth}
        \centering
        \includegraphics[width=\textwidth]{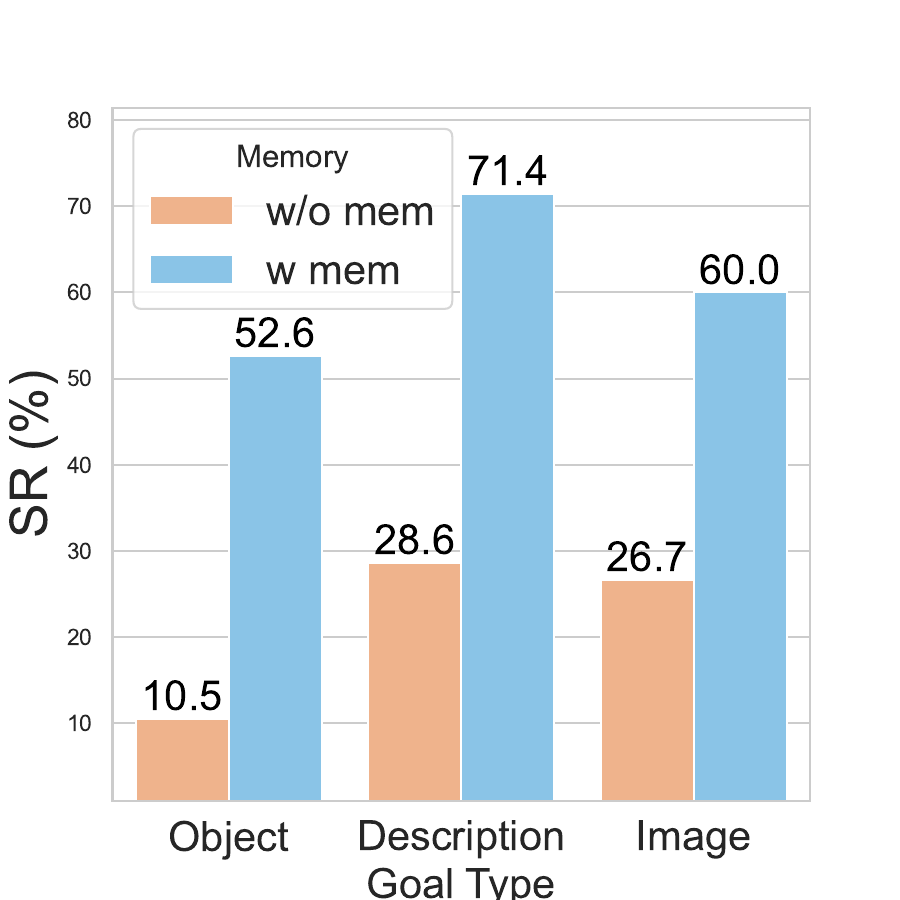}
        \caption{Usefulness of spatial memory.}
        \label{fig:memory-ablation}
    \end{subfigure}
    \hfill
    \begin{subfigure}[b]{0.45\textwidth}
        \centering
        \includegraphics[width=\textwidth]{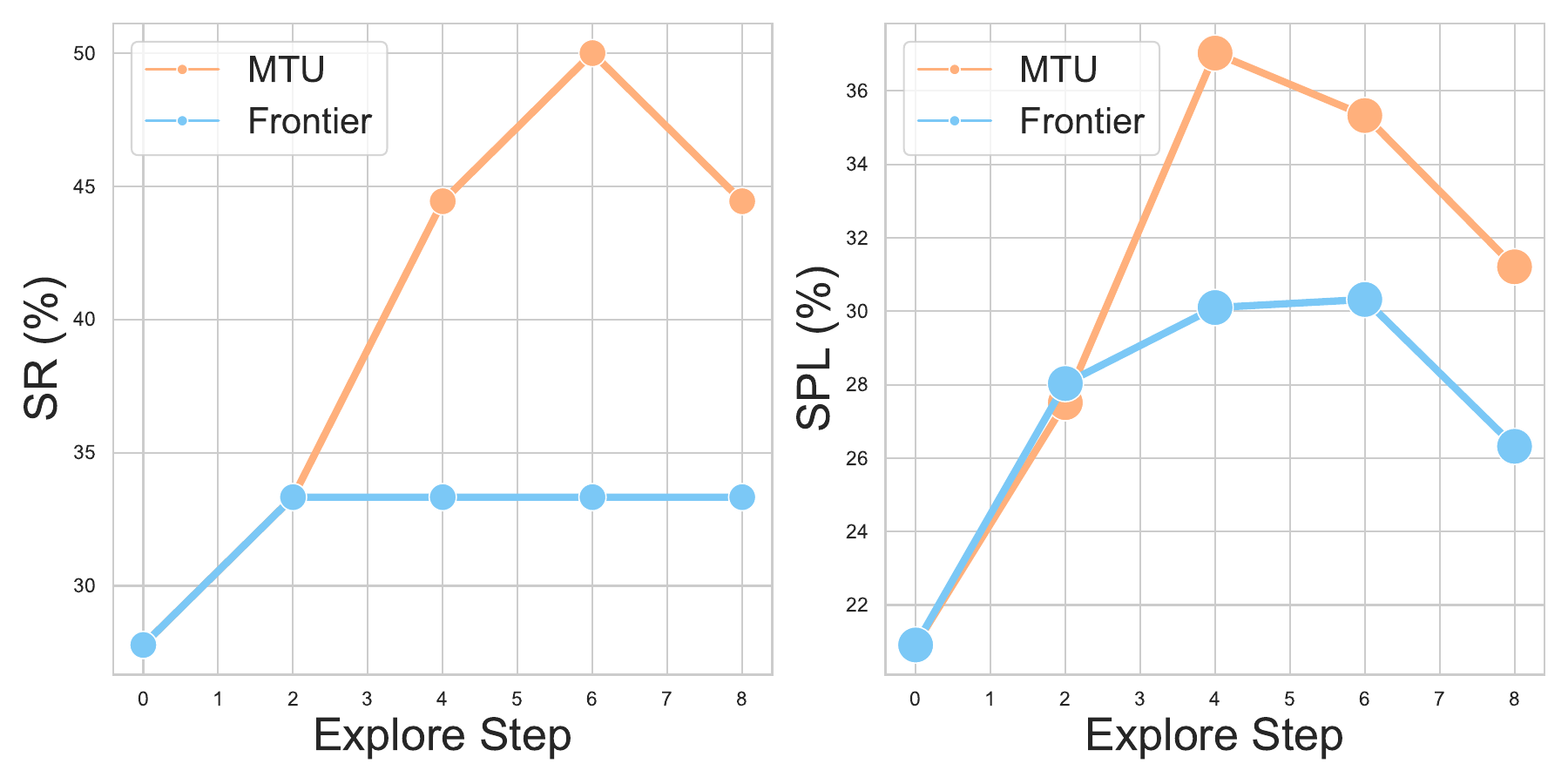}
        \caption{Exploration efficiency comparison. }
        \label{fig:explore-ablation}
    \end{subfigure}
    \caption{Ablation studies showing (a) the impact of vision-language-exploration pretraining, (b) exploration efficiency on seen environments, and (c) the contribution of spatial memory to navigation performance.}
    \label{fig:ablations}
\end{figure*}

\noindent \textbf{Does Vision-Langauge-Exploration Pe-training benefit navigation?}
The results in the \cref{fig:pretrain-ablation} show that Vision-Language Exploration (VLE) Pre-training significantly improves navigation performance, as indicated by the SR across all datasets. Specifically, SR increases from 27.8\% to 33.3\% in OVON, 22.2\% to 36.1\% in GOAT, and 22.9\% to 27.9\% in SG3D, demonstrating a consistent benefit of VLE across different task settings and distribution. 

\noindent \textbf{Does grounded training lead to efficient exploration?}
The results in \cref{fig:explore-ablation} demonstrate that MTU surpasses frontier exploration in both SR and SPL as exploration steps increase. Unlike frontier exploration, which blindly selects the nearest frontier, MTU utilizes semantic guidance for more efficient, goal-directed exploration. Notably, at step 6, MTU achieves a higher SR (50.0\% vs. 33.3\%) and improved SPL (35.3\% vs. 30.3\%).

\noindent \textbf{Can spatial memory bank enhance lifelong navigation?}
We reset spatial memory in w/o mem for each sub-episode in GOAT-Bench, and the experimental results in \cref{fig:memory-ablation} show that memory significantly improves SR across all goal types. Notably, SR increases from 10.5\% to 52.6\% for Object goals, 28.6\% to 71.4\% for Description goals, and 26.7\% to 60.0\% for Image goals, demonstrating that memory helps retain useful spatial information. 

\noindent \textbf{Can \model run in real time?}
\cref{tab:model-speed} shows that our model achieves efficient query proposal (192 ms) and fast reasoning (31 ms) while maintaining a competitive FPS (3.4) with 266M parameters. These results highlight the model's optimal balance between speed and performance, making it well-suited for real-time applications.
\begin{table}[h]
    \centering
    \renewcommand{\arraystretch}{1.2}
        \resizebox{0.85\columnwidth}{!}{
    \begin{tabular}{cccc}
        \toprule
        Query Proposal & Spatial Reasoning & FPS & Params \\
        \midrule
        192 ms & 31 ms & 3.4 & 266M \\
        \bottomrule
    \end{tabular}
    }
    \caption{Model speed and parameter metrics, results are average from 5 runs across multiple frames and episodes on 3090 Ti.}
    \label{tab:model-speed}
\end{table}

\subsection{Qualitative results}
\begin{figure}[!t]  
    \centering
    \includegraphics[width=0.5\textwidth]{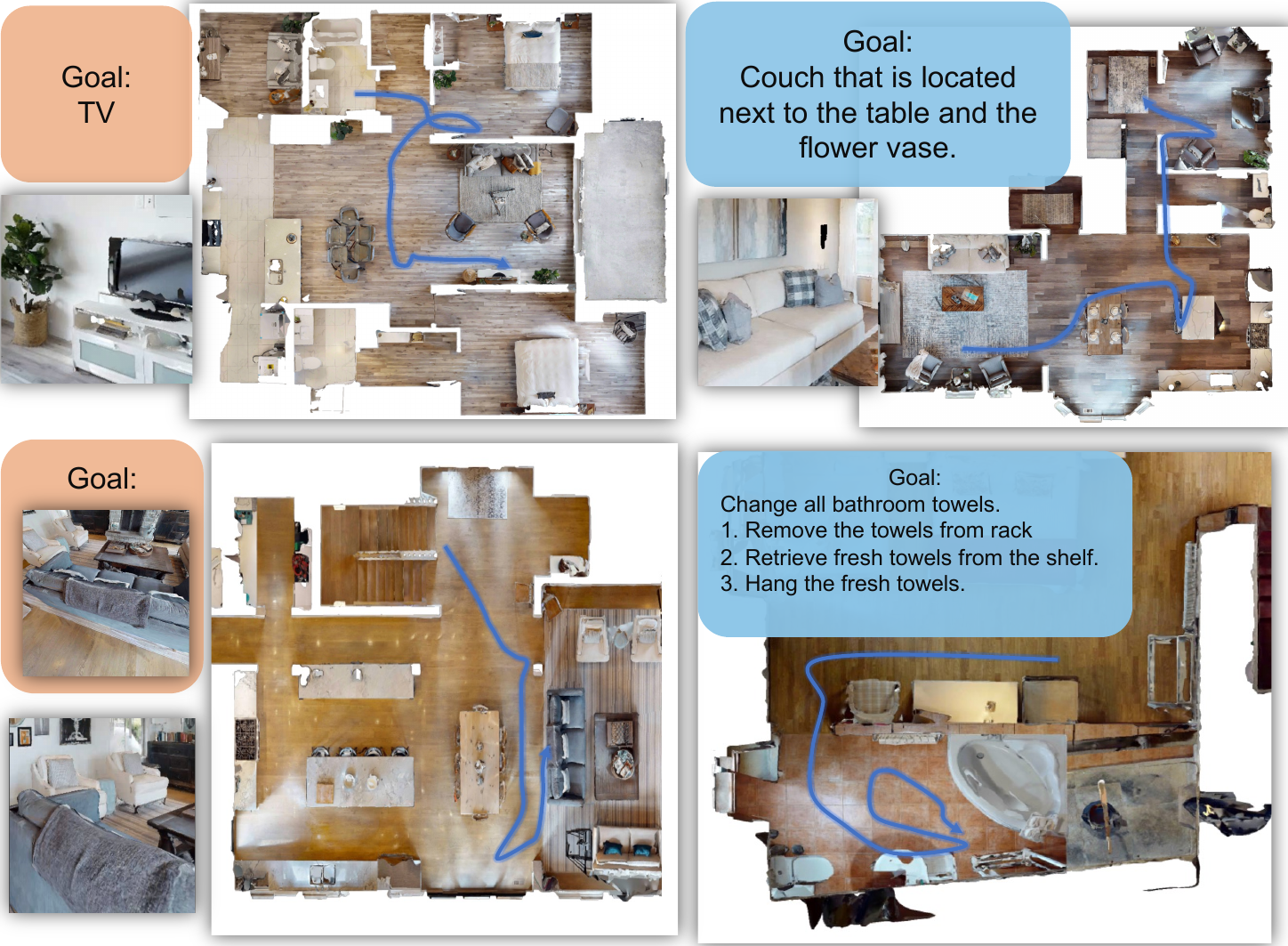} 
    \caption{Visualization of results in Habitat-Sim.} 
    \label{fig:qualitative} 
\end{figure}

These qualitative results in \cref{fig:qualitative} showcase the agent's ability to navigate and complete diverse goal types, including language, image, description, and task planning goals. The trajectories illustrate how the agent efficiently locates objects based on visual and semantic cues, demonstrating its capability to understand both image-based and textual instructions. Notably, for task planning goals, the agent follows a structured sequence of actions. 

\subsection{Real World Testing}
\begin{figure}[!t]  
    \centering
    \includegraphics[width=0.5\textwidth]{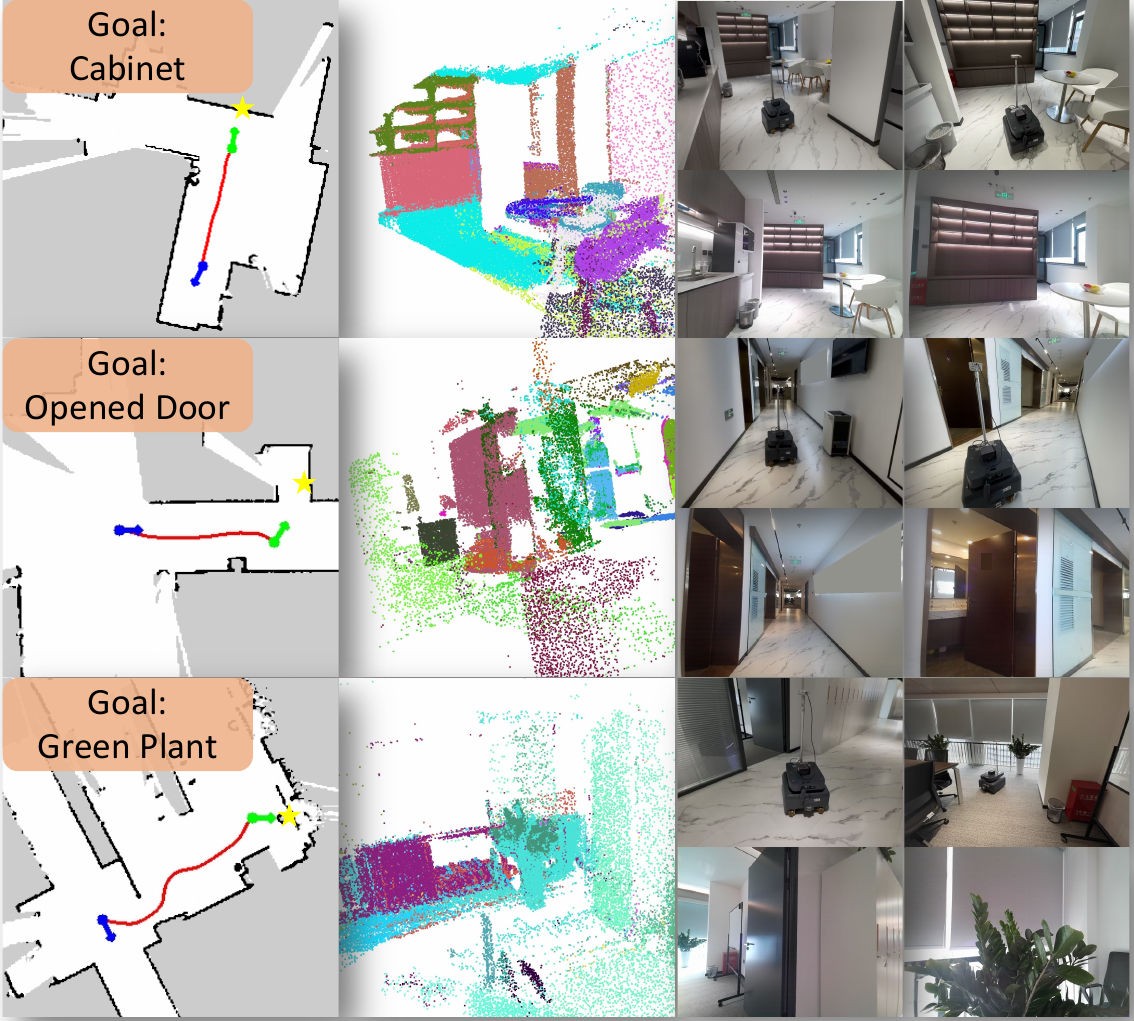} 
    \caption{Real World Testing.} 
    \label{fig:real_world_demo} 
    \vspace{-4mm}
\end{figure}

We evaluate MTU3D in realistic 3D environments by deploying it on an NVIDIA Jetson Orin with a Kinect~\cite{kinect} for real-time RGB-D data and a mobile robot equipped with Lidar for exploration. Without any real-world fine-tuning, we test the model in three diverse scenes: home, corridor, and meeting room. As shown in \cref{fig:real_world_demo}, MTU3D effectively navigates to the target position. Since MTU3D is trained on both simulated and real data, it overcomes the Sim-to-Real transfer challenges commonly faced in RL-based methods. This ability not only enhances its real-world applicability but also makes it highly scalable and impactful for advancing embodied intelligence in the future.

\label{sec:experiment}
\section{Conclusions}
In this paper, we introduce Move to Understand (\textbf{\model}), a unified framework that bridges visual grounding and exploration to advance embodied scene understanding. By jointly optimizing grounding and exploration, \model enables efficient navigation across diverse input modalities, fostering a deeper understanding of spatial environments. Our \textbf{V}ision-\textbf{L}anguage-\textbf{E}xploration (VLE) training leverages large-scale trajectories, achieving state-of-the-art performance on multiple Embodied AI benchmarks. Experimental results highlight the crucial role of \textit{spatial memory} in enabling lifelong multi-modal navigation and spatial intelligence, allowing agents to reason about and adapt to complex environments. Furthermore, real-world deployment demonstrates \model’s generalization ability, validating the effectiveness of our mixed training with both simulation and real-world data. By bridging visual grounding and exploration, \model paves the way for more capable, scalable, and generalizable embodied agents, bringing us closer to the goal of truly intelligent embodied AI.

\clearpage
\noindent\textbf{Acknowledgements.}
This work was supported in part by the National Science Foundation of China (NSFC) under Grant No. 62176134 and by a grant from the Assisted Medical Consultation Project Based on DeepSeek.

{
    \small
    \bibliographystyle{ieeenat_fullname}
    \bibliography{reference_header,main}
}


\appendix
\section{More Implementation Details}

\noindent \textbf{Local Query Refinement.} After obtaining the initial queries, we refine them through a series of decoder layers inspired by PQ3D. Specifically, within each decoder layer \( l \), our goal is to better retrieve object-relevant information by enhancing the interaction between the object queries \( Q_t^{l} \) and the input features \( \{\hat{F}_t^{2D}, \hat{F}_t^{3D}\} \). To achieve this, we first employ a cross-attention mechanism, allowing the object queries \( Q_t^{l} \) to attend to the input features \( \{\hat{F}_t^{2D}, \hat{F}_t^{3D}\} \). This step is crucial for aggregating relevant information from both 2D and 3D features.

To further improve the efficiency and performance of the cross-attention, we adopt a masked attention mechanism. This restricts the attention scope to localized features centered around each query, ensuring that the queries focus only on relevant regions rather than the entire feature map. This approach not only enhances the model's ability to capture fine-grained details but also significantly reduces computational overhead by limiting the attention span.

Following the cross-attention, we introduce a spatial self-attention module. This module leverages the coordinates of the queries to explore their spatial relationships, enabling the model to better understand the positional context of each query. By incorporating spatial information, the model can more effectively distinguish between different objects and their locations within the scene.

Each attention layer is followed by a forward feed network (FFN) and a normalization layer to stabilize the training process and enhance the representation learning capabilities of the model. The entire process within a single decoder layer can be formulated as follows:

\begin{align}
    &Q_t^{l'} = \text{FFN}\left(\text{Norm}\left(Q_t^{l} + \sum_{F \in F_{in}}\text{CrossAttn}(Q_t^{l}, F)\right)\right) \\
    &Q_t^{l+1} = \text{FFN}\left(\text{Norm}\left(\text{SpatialSelfAttn}(Q_t^{l'})\right)\right)
\end{align}

After \( L \) decoder layers, the refined local queries \( Q_t^{L} \) are expected to effectively capture the current observations. These refined queries encapsulate both the spatial and semantic information from the input features, making them highly representative of the objects within the scene.




\begin{figure*}[!t]
    \centering
    \captionsetup{type=figure}
    \includegraphics[width=0.97\textwidth]{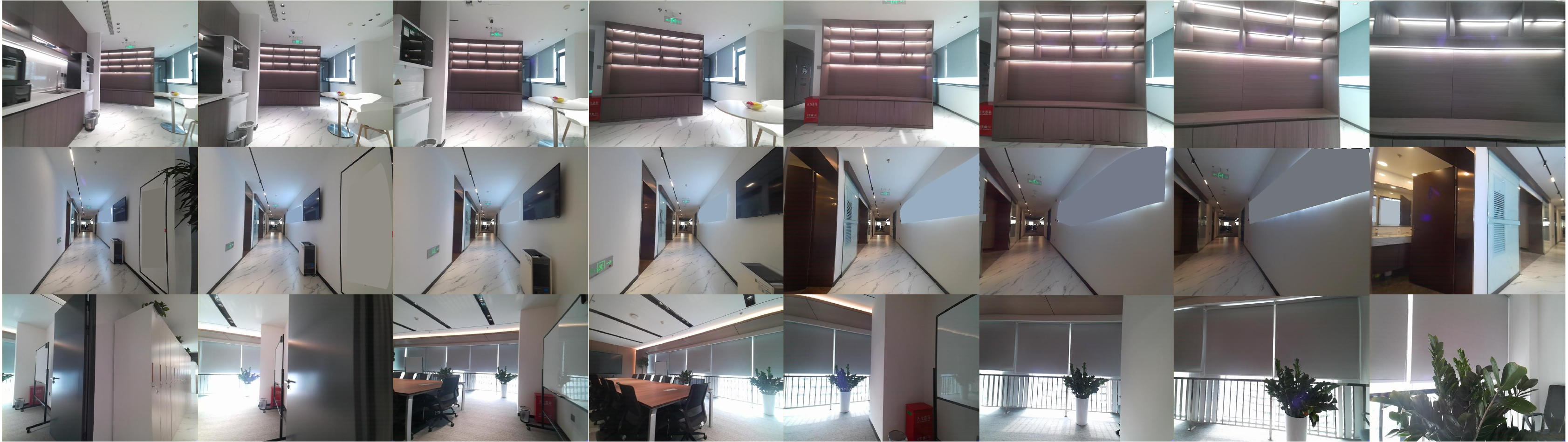}
    \captionof{figure}{Real world trajectory.}
    \label{fig:real_trajectory}
\end{figure*}

\noindent \textbf{Query Matching and Fusion.}
In our approach, we introduce a Dynamic Spatial Memory Bank to efficiently manage and update the queries across steps. Through extensive experiments, we observe that geometric similarity alone is often sufficient for matching queries. This observation motivates our design to leverage the geometric information of object queries to establish correspondences between the current queries and previous queries.

Given the current local queries \( Q_t^{L} \) and previous global queries \( Q_{t-1}^{G} \), we can obtain their respective bounding boxes: the current local bounding boxes \( B_t^{L} \in \mathbb{R}^{M \times 6} \) and the historical global bounding boxes \( B_{t-1}^{G} \in \mathbb{R}^{N \times 6} \). Here, \( M \) and \( N \) represent the number of local and global queries, respectively.

Our model is designed to predict global geometry based on partial input, thanks to a box loss function that aligns the predicted bounding boxes with ground-truth global values. This capability is crucial for establishing correspondences between the current and previous queries, even when only partial information is available.

To measure the similarity between the current local queries and previous global queries, we compute the Intersection over Union (IoU) matrix \( C \) between the bounding boxes \( B_t^{L} \) and \( B_{t-1}^{G} \):

\begin{equation}
    C = \text{IoU}(B_t^{L}, B_{t-1}^{G})
\end{equation}

Here, \(\text{IoU}(\cdot, \cdot)\) denotes the element-wise IoU score between two sets of axis-aligned bounding boxes. To ensure robust matching, we set elements in \( C \) that are smaller than a predefined threshold \( \epsilon \) to \( -\infty \). This step effectively filters out low-confidence matches and focuses on high-similarity pairs.

We then perform matching between \( B_t^{L} \) and \( B_{t-1}^{G} \) based on the cost matrix \( -C \). Each current local bounding box \( b_{t}^L \) is assigned to a previous global bounding box \( b_{t-1}^G \) if their similarity score is above the threshold. If a new local bounding box \( B_{t}^L[i] \) fails to match with any previous global bounding box \( b_{t-1}^G \), we register the corresponding query \( q_{t}^L[i] \) into the global queries as a new entry. This step ensures that newly detected objects are incorporated into the global representation.

For matched query pairs, we perform fusion to update the global representation. Specifically, instance segmentation masks \( m_t \) are fused through a union operation to maintain the most comprehensive segmentation information. For other representations, such as bounding box coordinates, we adopt a weighted average fusion strategy:

\begin{equation}
    B_{t}^G[i] = \frac{n}{n+1}B_{t-1}^G[i] + \frac{1}{n+1}B_{t}^L[j]
\end{equation}

Here, we assume that the \( j \)-th current local query is matched with the \( i \)-th previous global query. The variable \( n \) denotes the number of queries that have been merged into \( Q_{t-1}^{G}[i] \) so far. This weighted average approach ensures that the global representation is updated smoothly, incorporating new information while retaining historical context.


\noindent \textbf{Spatial Reasoning.}
The Spatial Reasoning Transformer is designed to integrate spatial context and language instructions for effective reasoning. It shares a similar transformer architecture with the local query refinement module but includes additional mechanisms to incorporate language goals. Specifically, the transformer operates on concatenated global and frontier queries, denoted as \( Q_t^G \) and \( Q_t^F \) respectively, to form the initial queries \( Q_t^0 \):

\begin{align}
    Q_t^0 &= \text{Concat}(Q_t^G, Q_t^F)
\end{align}

For each decoder layer \( l \), the queries first attend to the input features \( F_{in} \) to aggregate spatial information from the current observation. This is achieved through a cross-attention mechanism followed by normalization and a feed-forward network (FFN):

\begin{align}
    Q_t^{l'} &= \text{FFN}\left(\text{Norm}\left(Q_t^{l} + \sum_{F \in F_{in}}\text{CrossAttn}(Q_t^{l}, F)\right)\right)
\end{align}

\begin{figure}[!t]  
    \centering
    \includegraphics[width=0.5\textwidth]{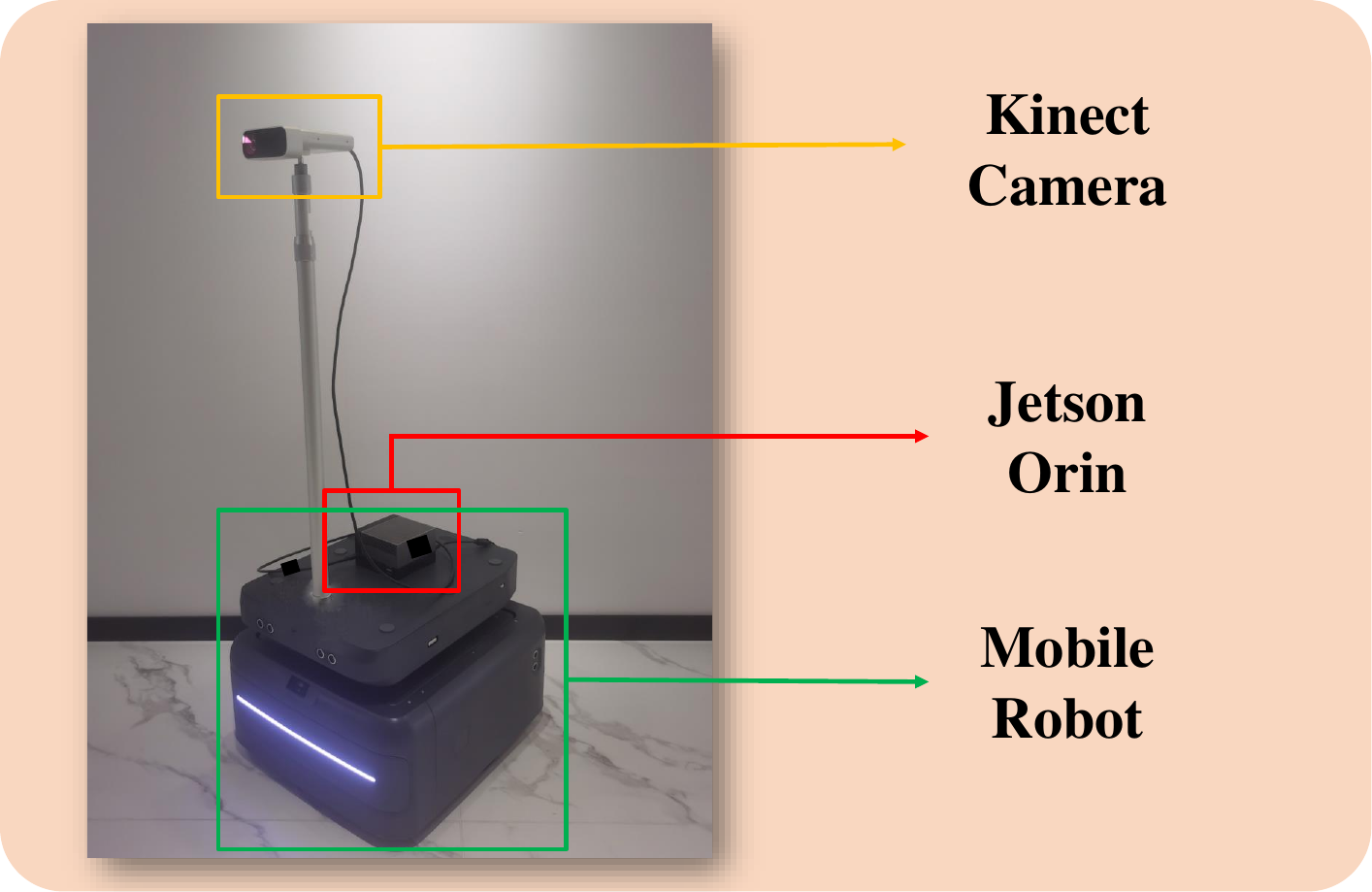} 
    \caption{Real Device.} 
    \label{fig:real_device} 
    \vspace{-4mm}
\end{figure}

Next, an additional cross-attention layer is introduced to incorporate the language goal \( L \). This step allows the model to align its spatial reasoning with the provided language instructions:

\begin{align}
    Q_t^{l''} &= \text{FFN}\left(\text{Norm}\left(Q_t^{l'} + \text{CrossAttn}(Q_t^{l'}, L)\right)\right)
\end{align}

Finally, a spatial self-attention layer is applied to capture the spatial relationships among the queries, further refining their representations:

\begin{align}
    Q_t^{l+1} &= \text{FFN}\left(\text{Norm}\left(\text{SpatialSelfAttn}(Q_t^{l''})\right)\right)
\end{align}

Through the process, the Spatial Reasoning Transformer effectively fuses spatial and linguistic information for further exploration decision.

\section{Benchmarks and baseline}
\noindent \textbf{HM3D-OVON.}
HM3D-OVON is an open-vocabulary navigation benchmark. Due to its large-scale test set and resource limitations, we random sample 360 episodes for evaluation. For baselines, BC trains the agent using supervised learning on expert trajectories. DAgger involves an expert providing corrective actions online during training. RL, BCRL, and DAgRL represent reinforcement learning from scratch, reinforcement learning initialized with behavior cloning, and reinforcement learning with DAgger, respectively. Uni-Navid is a video-based navigation method, while TANGO is a training-free navigation approach that leverages large language models.

\noindent \textbf{Goat-Bench.}
Goat-Bench evaluate multi-modal life long navigation, goals include image, class, and description, due to the large-scale test set and resource limitations, we random sample 90 tasks for evaluation. Modular Goat and Modular clip one wheels represent module approaches using pre-trained detector and feature for zero-shot navigation. SenseAct-NN Skill Chain and SenseAct-NN Monotholic are RL approaches, with single head or multi head for each goat type. 

\noindent \textbf{SG3D.}
Sequential Navigation requires an agent to navigate to a target object in a specified order within a 3D simulation environment. The Embodied Video Agent is a modular approach that incorporates persistent memory and utilizes a large language model  as the planner. SenseAct-NN Monolithic is identical to the variant used in Goat-Bench, employing a unified reinforcement learning policy for all goal types.

\noindent \textbf{A-EQA.}
A-EQA evaluates a model’s ability to explore an environment in response to a given question. For A-EQA evaluation, our model is solely responsible for generating the exploration trajectory and collecting the corresponding video for each question. The question answering itself is handled by GPT-4o/V. To ensure a fair comparison, we use the same prompts and the same number of video frames as the baseline methods when measuring exploration performance.

\section{Trajectory collection}
Algorithm 1 outlines our trajectory collection strategy, in which we randomly select each action to simulate the behavior of agents in HM3D. Relying solely on random or ground-truth actions can lead to model overfitting.
\begin{algorithm}[!t]
\caption{Explore an Episode}

\textbf{Global}: \textit{explored\_map}, \textit{visited\_frontiers}, \textit{visible\_ids}

\textbf{END type}: \textbf{Success}, \textbf{Unreachable}, \textbf{Invisible}, \textbf{Failure}
\begin{algorithmic}[1]


\Function{Explore\_an\_Episode}{strategy, goals}
    \State \textit{decision\_list} $\gets$ []
    \While{not END}
    
        \State Spin and Update
        
        \State \textit{goal} $\gets$ select closest goal in \textit{goals}
        
        \State \textit{visible} $\gets$ \textit{goal} \textbf{in} \textit{visible\_ids}
        
        \State \text{reachable} $\gets$ \textit{goal} \textbf{in} \textit{explored\_map}
        
        
        
        \State \textit{better\_chance} $\gets$ \textbf{exist} frontier closer to \textit{goal}
        
        \If{\textit{visible} and \textit{reachable}}
            \State Record decision and Goto \textit{goal}, \textbf{Success}
        \ElsIf{\textit{better\_chance}}
            \State Record decision and Goto next frontier
        \ElsIf{\textit{visible} but \textbf{not} \textit{reachable}}
            \State \textbf{Unreachable}
        \ElsIf{ \textit{reachable} but \textbf{not} \textit{visible}}
            \State \textbf{Invisible}
        \ElsIf{\textbf{not} \textit{visible} and \textbf{not} \textit{reachable}}
            \State \textbf{Failure}
        \EndIf

        
    \EndWhile
    \State \textit{status} = \textbf{END type}
    \State \Return (\textit{decision\_list}, \textit{status})
\EndFunction
\end{algorithmic}
\end{algorithm}

\end{document}